%% file: main.tex
\documentclass{article} 
\usepackage{iclr2025_conference,times}
\usepackage{mathtools}
\usepackage{tikz}
\usepackage[list=true]{subcaption} 
\input{math_commands.tex}

\usepackage{hyperref}
\usepackage{url}
\usepackage{soul}
\usepackage{booktabs}
\usepackage{makecell}
\usepackage{multirow}

\usepackage{wrapfig}
\definecolor{almond}{rgb}{0.94, 0.87, 0.8}
\definecolor{antiquewhite}{rgb}{0.98, 0.92, 0.84}
\definecolor{electriclavender}{rgb}{0.96, 0.73, 1.0}
\definecolor{rdd}{rgb}{1, 0.6, 1} 
\definecolor{resultcolor}{rgb}{1, 0.8, 1} 

\newcommand\mydots{\hbox to 1em{.\hss.\hss.}}

\newcommand{\Real}[1]{\mathbb{R}^{#1}}
\newcommand{\Rl}[2]{\mathbb{R}^{#1 \times #2}}

\renewcommand{\vec}{\mathsf{vec}}

\sethlcolor{magenta}

\title{Mimetic Initialization of MLPs}

\author{Asher Trockman\thanks{Asher is now at Google. Work done at CMU.},\; J. Zico Kolter \\
Carnegie Mellon University \\
Correspondence to: \texttt{ashert@cs.cmu.edu}
}

\iclrfinalcopy 
\begin{document}

\maketitle

\begin{abstract}
Mimetic initialization uses pretrained models as case studies of good initialization,
using observations of structures in trained weights to inspire new, simple
initialization techniques.
So far, it has been applied only to spatial mixing layers, such as convolutional, self-attention,
and state space layers.
In this work, we present the first attempt to apply the method to channel mixing layers,
namely multilayer perceptrons (MLPs).
Our extremely simple technique for MLPs---to give the first layer a nonzero mean---speeds up training
on small-scale vision tasks like CIFAR-10 and ImageNet-1k.
Though its effect is much smaller than spatial mixing initializations, it
can be used in conjunction with them for an additional positive effect.
\end{abstract}

\section{Introduction}

\paragraph{Background}
Neural network weights are typically initialized at random
from univariate distributions, as in Xavier~\cite{xavier}
and Kaiming~\cite{kaiming} initializations.
These techniques focus on the \emph{signal propagation} perspective,
where the variance of the univariate random weights is scaled according
to the dimensions of the layer weights
with the goal of preventing vanishing and exploding gradients.

However, modern deep networks use normalization layers
such as BatchNorm~\citep{batchnorm} and LayerNorm~\citep{layernorm}
in addition to adaptive optimizers like Adam~\citep{adam};
taken together, these methods make precise control of scale at initialization
somewhat less critical. Of course, scale still plays an important
role~\citep{kunin}.

The most effective initialization is \emph{pretraining}, or transfer learning~\citep{bit};
pretrained networks have stored transferrable knowledge
that can be leveraged to quickly adapt to downstream tasks.
Consequently, pretrained networks are much easier to train.
But could it be that this transferrable knowledge is only part of the story---
that the geometry of the weight space of pretrained networks
simply makes them easier to optimize, in a way we could capture or study
without pre-training?

\paragraph{Related work}
Recently, several works on \emph{mimetic initialization} have proposed
that the effect of pretraining can be decomposed into two components:
(1) \emph{storing knowledge} and (2) \emph{serving as a good initialization}~\citep{lego,convcov,mimetic,ssm,alex}.
Moreover, it seems that to some extent, the hypothesized \emph{good initialization}
component can be disentangled from the other by ``hand'', in closed form.
Mimetic initialization uses pretrained models
as case studies of good initialization, and summarizes the weight space
structures observed in simple but \emph{multivariate} initialization schemes
that specify the high-dimensional statistical dependence between weights.

\looseness=-1
A mimetic initialization for convolutions~\citep{convcov} specified the structure of
weights for individual filters depending on the depth of the layer.
Mimetic initialization for self-attention layers~\citep{mimetic} attempted to make self-attention
more ``convolutional'' (localized receptive field) at initialization, among other
observations: the query and key weights should be correlated, value and projection
weights anti-correlated, and sinusoidal position embeddings should be used at initialization;
see also~\cite{zheng2024convolutional}.
Most recently, mimetic initialization for state space layers~\citep{ssm} used
the correspondence between state space layers and linear self-attention layers to leverage
the previous technique.
These methods allow training networks in fewer steps and to higher final
accuracies, especially in small-scale and data-limited settings.

\paragraph{This work} All previous work on mimetic initialization
has focused on \emph{spatial mixing} layers, i.e., large-filter convolutions,
self-attention layers, and state space layers. In this work, we investigate whether
the same technique may apply to some \emph{channel mixing} layers, in particular
MLPs, one of the most important primitives in deep learning, i.e.,
y = $W_2 \sigma(W_1x)$ for nonlinear activation function $\sigma$ and
projections $W_1 \in \Rl{p}{n}$, $W_2 \in \Rl{n}{p}$ and $x \in \Real{n}$.
The importance of the initialization of the value and projection weights in
\citet{mimetic} could be viewed as an application of mimetic init to channel mixing layers;
however, this finding only held in conjunction with the attention map (query/key) init itself.
In contrast, we focus on MLPs in isolation.
While we observe interesting statistical structure in trained MLPs,
our only empirically-backed finding so far suggests a modification to the mean of $W_1$.

\section{Understanding the covariance structure of trained MLPs}

\begin{figure}
    \centering
   \begin{subfigure}[t]{0.495\linewidth}
       \includegraphics[width=\linewidth]{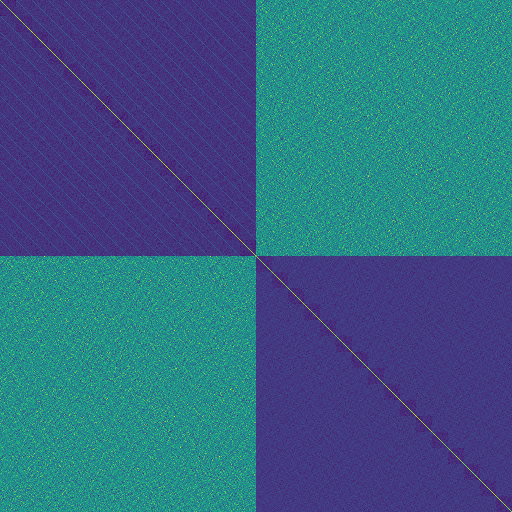}
        \caption{ConvNeXt}
        \label{fig:cov-convnext}
    \end{subfigure}%
    \hfill%
    \begin{subfigure}[t]{0.495\linewidth}
        \includegraphics[width=\linewidth]{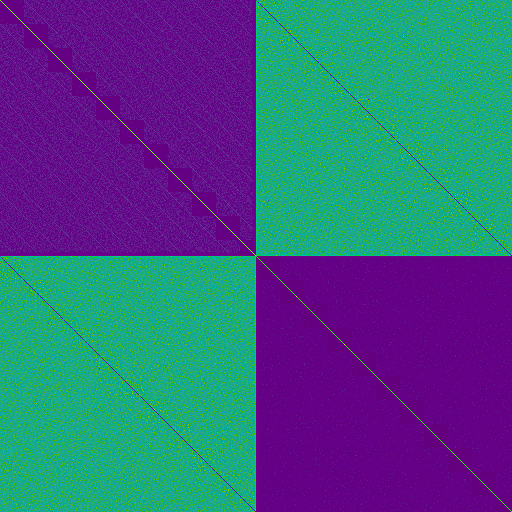}
        \caption{ViT (enhanced; may need to zoom for structure)}
        \label{fig:cov-picovit}
    \end{subfigure}
   \caption{Full empirical covariance matrices for unrolled MLP weights $W_1, W_2$ (jointly).}
\end{figure}

Previous work on mimetic initialization has focused on the covariance structure
of pretrained weights.
\cite{convcov} studied the covariance structure of convolutional filters \emph{within single networks}, i.e., a network with embedding dimension $n$ has $n$ filter samples per $l$
layers, from which one can calculate $l$ empirical covariance matrices.
\cite{mimetic} study the cross-covariance of query/key and value/projection weights,
again focusing on samples of rows or columns within individual networks.

In this work, we adopt a somewhat more powerful technique; instead of studying
empirical covariances calculated \emph{within single networks} over slices
of weights (rows, columns, filters, etc.), we study empirical covariances calculated
over \emph{populations of networks}. That is, we study the distribution of
$[\vec(W_1)\; \vec(W_2)] \in \Real{2np}$ rather than, say, $[W_1[i]\; W_2[:,i]] \in \Real{2n}$.

We start our investigation by training tiny ConvNeXt~\citep{convnext} models,
a simple convolutional architecture similar to ConvMixer~\citep{convmixer}
that uses isolated MLP layers (alternately with depthwise convolution) instead of linear layers.
We train $\approx 10^4$ such models on CIFAR-10 classification to estimate weight space covariances;
as this is computationally expensive, we use models with only a few thousand
or up to $\approx 10^4$ parameters. 
For example, we use just 3--4 layers and an internal dimension of around 16--32
with MLP expansion factor 2. We train over thousands of different seeds
and unroll the MLP weights $W_1, W_2$ to compute the full empirical covariances
of each weight $\mathsf{Cov}(\vec\, W)$ and cross covariance $\mathsf{Cov}(\vec\, W_1,\vec\,  W_2)$.

In Fig.~\ref{fig:cov-convnext}, we show the full empirical covariance matrix of
$[\vec(W_1)\; \vec(W_2)]$. Notably, the covariances corresponding to each weight matrix
are striped. This suggests that the rows or columns (depending on row/column major order for unrolling) are correlated in trained MLPs. For example, each row or column could have
a nonzero ``group mean'' while the overall weight matrix still has a global/``grand'' mean
of zero. This group mean could itself be randomly distributed with mean zero.
We also note that the stripes tend to be more pronounced in $\mathsf{Cov}(W_1)$ than in $\mathsf{Cov}(W_2)$.

We attempt to mimic this very simple structure at initialization time.
We add a small amount of random noise to $b_n\sim \mathcal{N}(0, \sigma I_n)$ to $W_1$,
broadcasting over rows:
\begin{equation}
    W_1' = W_1 + \mathbf{1}_pb_n^T.\label{eq:init1}
\end{equation}
This mimics the structure seen in pretrained weights, and we find that it also speeds up training
considerably.
However, we note that an even simpler solution exists to mimic the structure in
Fig.~\ref{fig:cov-convnext}: simply add a single, constant small bias term $b \in \Real{}$ to $W_1$:
\begin{equation}
    W_1' = W_1 + b \mathbf{1}_p \mathbf{1}_n^T. \label{eq:init2}
\end{equation}
Both strategies result in randomly-sampled matrices with significant stripes as in Fig.~\ref{fig:cov-convnext}.
In our experiments, this extremely simple tweak is surprisingly effective at
increasing the trainability of small ConvNeXts.

In Fig.~\ref{fig:cov-picovit}, we do the same analysis for a small Vision Transformer (ViT)
trained in the same setup. We see the same patterns as in Fig.~\ref{fig:cov-convnext};
moreover, we see that $W_1$ and $W_2$ weights are apparently anti-correlated
as noted for the value and projection weights in previous work~\citep{mimetic};
that is, a mimicking initialization may be:
$W_1' = \tfrac{1}{2}(W_1 - W_2).$
We do not see this pattern within ConvNeXt. Further, we saw no benefit
to initializing MLP weights to be anticorrelated, unlike our $W_1$-mean initialization
above.

\section{Experiments}

\begin{figure}
    \centering
   \begin{subfigure}[t]{0.495\linewidth}
       \includegraphics[width=\linewidth]{"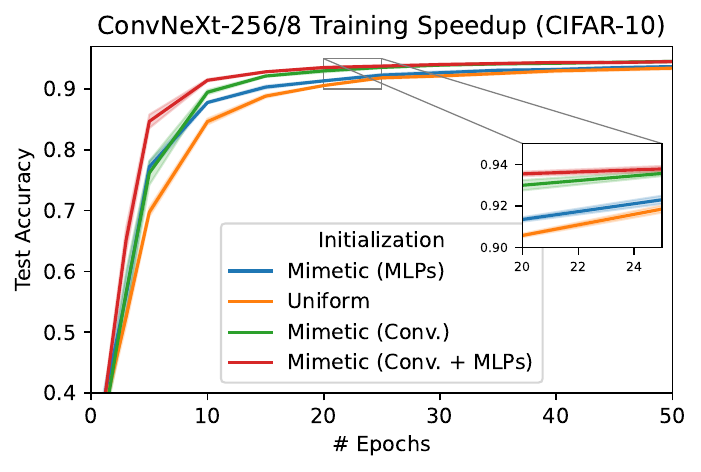"}
    \end{subfigure}%
    \hfill%
    \begin{subfigure}[t]{0.495\linewidth}
        \includegraphics[width=\linewidth]{"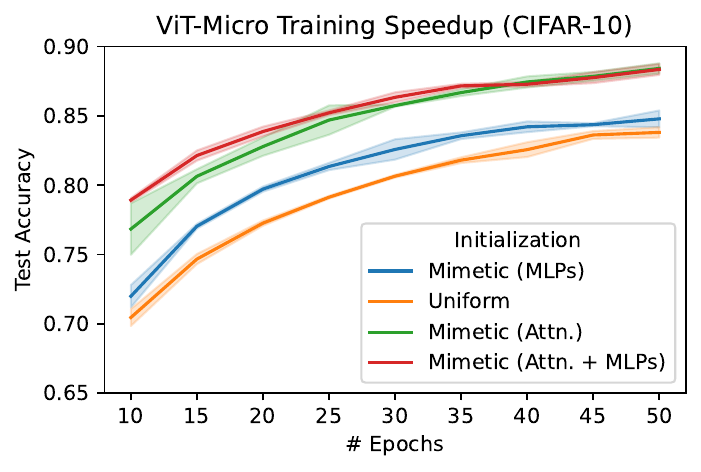"}
    \end{subfigure}
   \caption{CIFAR-10 experiments for ConvNeXt and ViT in conjunction with previous
   mimetic inits for convolutional and self-attention layers. The effect of the MLP init is significant in early training, but eventually evens out. The effect of conv. and attn. mimetic init, however, remains. \emph{Note:} each point in the graph represents
   the accuracy of a completed training run of $x$ epochs, mean/std. reported over 5 seeds.}
   \label{fig:convnext}
\end{figure}

\begin{figure}
\centering
\begin{minipage}{.48\textwidth}
  \centering
  \includegraphics[width=\textwidth]{"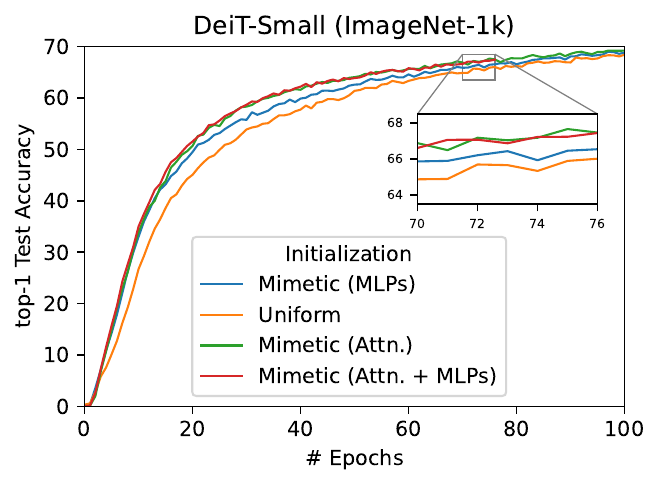"}
  \caption{Our MLP init improves early ImageNet-1k training
  in a data-efficient ViT, and maintains a 0.5\% advantage
  over baseline for the full training time.}
  \label{fig:imnet}
\end{minipage}\hfill
\begin{minipage}{.48\textwidth}
  \centering
  \includegraphics[width=\textwidth]{"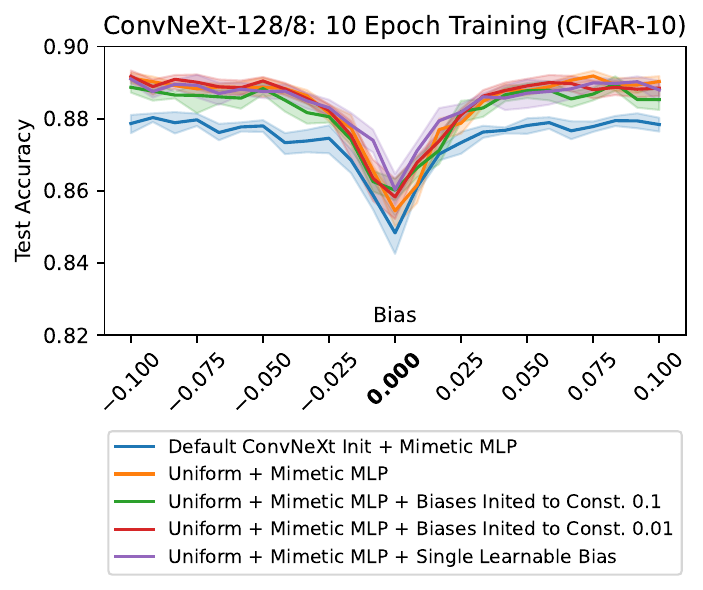"}
  \caption{Sweep of the bias parameter $b$ of our init;
  across several baselines, $b=0$ is noticeably suboptimal. Mean/std. over 5 seeds per $b$.}
  \label{fig:bias-sweep}
\end{minipage}
\end{figure}

We test our technique on small ConvNeXt and ViT models trained on CIFAR-10
with RandAugment, using the same pipeline as in~\citet{convcov}.
We made some noteworthy changes to the default initialization scheme of ConvNeXt,
beyond our proposed technique: we replaced the truncated normal initialization
with fixed standard deviation $\sigma=0.02$ and zero bias initialization
with the standard Kaiming uniform initialization in PyTorch;
this lead to broadly improved accuracies for our training setup.
The CIFAR-10 models presented use $2\times2$ patches for $32\times32$ inputs,
and the ImageNet models $16\times16$ for $224\times224$ inputs.

We find that the simple initialization tweak in Eq.~\ref{eq:init2}
improves ConvNeXt and ViT results for small scale and short-duration training.
In Fig.~\ref{fig:bias-sweep}, we show accuracy for a variety of settings of $b$;
a wide variety of settings improve performance, with a notable dip at $b=0$.
However, unlike previous work on mimetic initialization, it seems that advantage
of using our proposed MLP init decreases with increasing training time (see Fig.~\ref{fig:convnext}).

\looseness=-1
\paragraph{ConvNeXt results}
In Fig.~\ref{fig:convnext}, we show the effect of training with various
mimetic init techniques compared to uniform (default) initialization,
on an isotropic ConvNeXt-256/8 with $7\times7$ filters.
Our MLP init significantly increases accuracy for short-duration training (such as 10-20 epochs), though the effect tapers off for longer training times (such as 50).
The MLP init also works with the convolution init
from~\cite{convcov}, together outperforming either mimetic init alone;
but in contrast, 
the convolution init maintains an advantage over uniform even after many epochs.
\paragraph{ViT results}
We also trained tiny ViTs with mimetic initialization,
which have dimension 192, depth 8, and 3 heads; see Fig.~\ref{fig:convnext}.
The benefit of our MLP init persists for longer training times
compared to ConvNeXt, but the additive effect
with self-attention mimetic init may be smaller.
In Fig.~\ref{fig:imnet}, we trained a DeiT-Small (22M params) on ImageNet-1k using
the training pipeline from~\citet{deit}.
Our init results in considerable initial gains in accuracy,
though the difference narrows significantly with training time;
using our MLP init \emph{with} self-attention mimetic init
results in rather small gains.
The difference may be larger in a
simpler ResNet-style training pipeline, as in~\cite{mimetic}.
Nonetheless, our method results in a consistent advantage of 0.5\%
over 100 epochs, achieving 68.8\% over the course of
100 epochs, compared to 68.4\% for the baseline.
\looseness=-1
\paragraph{Is it really the init?} 
We
thought our method may just be compensating for the default linear layer bias
of zero in the ConvNeXt implementation, but this was not the case---
see Fig.~\ref{fig:bias-sweep}.
One baseline
is to simply add a learnable scalar bias per $W_1$,
as $b \mathbf{1}_p \mathbf{1}_n^T x = b \cdot \mathsf{sum}(x) \cdot \mathbf{1}_p$.
This does not absorb
the effect of our init (see Fig.~\ref{fig:bias-sweep}).
We also attempted to change the init of the bias of the linear layer itself
to a small constant.
However, no baselines explain the effect of our MLP init.


\section{Conclusion}

We presented a simple concept for a mimetic initialization for MLPs;
our method is to simply initialize the first layer of MLPs
with a small nonzero mean.
We have a small amount of preliminary 
evidence to suggest that using Eq.~\ref{eq:init1} instead of Eq.~\ref{eq:init2} 
for larger-scale training may be better;
Eq.~\ref{eq:init1} maintains the global mean of zero and
may encode a weaker inductive bias than Eq.~\ref{eq:init2}.
However, more research at larger scale is necessary to draw conclusions.
While our results are not as robust as those in the mimetic 
initialization works for spatial mixing layers, we think our finding is
nonetheless interesting in terms of understanding the ``good initialization''
component of pretraining. We also suspect that, like previous work,
our technique would be most important in data-limited settings.

\looseness=-1
We tried to exploit one of the simplest structures we have seen
in the weight space, but many others exist, 
such as the anticorrelation of weights in ViT MLPs.
More generally, we think that our method
of studying populations of pretrained network weights, as opposed to
samples of weights within single networks, could lead to further insights
in the growing field of weight space learning and analysis.

\bibliography{iclr2025_conference}
\bibliographystyle{iclr2025_conference}


\end{document}

%% file: math_commands.tex
\usepackage{amsmath,amsfonts,bm}

\def\eqref#1{equation~\ref{#1}}

\def\1{\bm{1}}

\DeclareMathAlphabet{\mathsfit}{\encodingdefault}{\sfdefault}{m}{sl}
\SetMathAlphabet{\mathsfit}{bold}{\encodingdefault}{\sfdefault}{bx}{n}



%% file: iclr2025_conference.bib
@inproceedings{mimetic,
  title={Mimetic initialization of self-attention layers},
  author={Trockman, Asher and Kolter, J Zico},
  booktitle={International Conference on Machine Learning},
  pages={34456--34468},
  year={2023},
  organization={PMLR}
}

@article{convcov,
  title={Understanding the covariance structure of convolutional filters},
  author={Trockman, Asher and Willmott, Devin and Kolter, J Zico},
  journal={arXiv preprint arXiv:2210.03651},
  year={2022}
}

@inproceedings{kaiming,
  title={Deep residual learning for image recognition},
  author={He, Kaiming and Zhang, Xiangyu and Ren, Shaoqing and Sun, Jian},
  booktitle={Proceedings of the IEEE conference on computer vision and pattern recognition},
  pages={770--778},
  year={2016}
}

@inproceedings{xavier,
  title={Understanding the difficulty of training deep feedforward neural networks},
  author={Glorot, Xavier and Bengio, Yoshua},
  booktitle={Proceedings of the thirteenth international conference on artificial intelligence and statistics},
  pages={249--256},
  year={2010},
  organization={JMLR Workshop and Conference Proceedings}
}

@article{ssm,
  title={Mimetic Initialization Helps State Space Models Learn to Recall},
  author={Trockman, Asher and Harutyunyan, Hrayr and Kolter, J Zico and Kumar, Sanjiv and Bhojanapalli, Srinadh},
  journal={arXiv preprint arXiv:2410.11135},
  year={2024}
}

@article{adam,
  title={Adam: A method for stochastic optimization},
  author={Kingma, Diederik P},
  journal={arXiv preprint arXiv:1412.6980},
  year={2014}
}

@article{batchnorm,
  title={Batch normalization: Accelerating deep network training by reducing internal covariate shift},
  author={Ioffe, Sergey},
  journal={arXiv preprint arXiv:1502.03167},
  year={2015}
}

@article{layernorm,
  title={Layer normalization},
  author={Lei Ba, Jimmy and Kiros, Jamie Ryan and Hinton, Geoffrey E},
  journal={ArXiv e-prints},
  pages={arXiv--1607},
  year={2016}
}

@article{zheng2024convolutional,
  title={Convolutional Initialization for Data-Efficient Vision Transformers},
  author={Zheng, Jianqiao and Li, Xueqian and Lucey, Simon},
  journal={arXiv preprint arXiv:2401.12511},
  year={2024}
}

@article{lego,
  title={Unveiling transformers with lego: a synthetic reasoning task},
  author={Zhang, Yi and Backurs, Arturs and Bubeck, S{\'e}bastien and Eldan, Ronen and Gunasekar, Suriya and Wagner, Tal},
  journal={arXiv preprint arXiv:2206.04301},
  year={2022}
}

@article{alex,
  title={On the Surprising Effectiveness of Attention Transfer for Vision Transformers},
  author={Li, Alexander C and Tian, Yuandong and Chen, Beidi and Pathak, Deepak and Chen, Xinlei},
  journal={arXiv preprint arXiv:2411.09702},
  year={2024}
}

@inproceedings{deit,
  title={Training data-efficient image transformers \& distillation through attention},
  author={Touvron, Hugo and Cord, Matthieu and Douze, Matthijs and Massa, Francisco and Sablayrolles, Alexandre and J{\'e}gou, Herv{\'e}},
  booktitle={International conference on machine learning},
  pages={10347--10357},
  year={2021},
  organization={PMLR}
}

@article{kunin,
  title={Get rich quick: exact solutions reveal how unbalanced initializations promote rapid feature learning},
  author={Kunin, Daniel and Ravent{\'o}s, Allan and Domin{\'e}, Cl{\'e}mentine and Chen, Feng and Klindt, David and Saxe, Andrew and Ganguli, Surya},
  journal={arXiv preprint arXiv:2406.06158},
  year={2024}
}

@inproceedings{bit,
  title={Big transfer (bit): General visual representation learning},
  author={Kolesnikov, Alexander and Beyer, Lucas and Zhai, Xiaohua and Puigcerver, Joan and Yung, Jessica and Gelly, Sylvain and Houlsby, Neil},
  booktitle={Computer Vision--ECCV 2020: 16th European Conference, Glasgow, UK, August 23--28, 2020, Proceedings, Part V 16},
  pages={491--507},
  year={2020},
  organization={Springer}
}

@article{convmixer,
  title={Patches are all you need?},
  author={Trockman, Asher and Kolter, J Zico},
  journal={arXiv preprint arXiv:2201.09792},
  year={2022}
}

@inproceedings{convnext,
  title={A convnet for the 2020s},
  author={Liu, Zhuang and Mao, Hanzi and Wu, Chao-Yuan and Feichtenhofer, Christoph and Darrell, Trevor and Xie, Saining},
  booktitle={Proceedings of the IEEE/CVF conference on computer vision and pattern recognition},
  pages={11976--11986},
  year={2022}
}
